\documentclass[letterpaper]{article} 
\usepackage{aaai25}  
\usepackage{times}  
\usepackage{helvet}  
\usepackage{courier}  
\usepackage[hyphens]{url}  
\usepackage{graphicx} 
\usepackage{natbib}  
\usepackage{caption} 
\usepackage{algorithm}
\usepackage{algpseudocode}
\usepackage{booktabs}
\usepackage{multirow}
\usepackage{amsmath}
\usepackage{amssymb}
\usepackage{subcaption}
\usepackage[page]{appendix} 
\usepackage{lipsum}

\usepackage{newfloat}
\usepackage{listings}
\DeclareCaptionStyle{ruled}{labelfont=normalfont,labelsep=colon,strut=off} 
\floatstyle{ruled}
\newfloat{listing}{tb}{lst}{}
\floatname{listing}{Listing}
\usepackage{cleveref}
\title{FLAME: Learning to Navigate with Multimodal LLM in Urban Environments}

\author{
    Yunzhe Xu\textsuperscript{\rm 1},
    Yiyuan Pan\textsuperscript{\rm 2},
    Zhe Liu\textsuperscript{\rm 1}\thanks{Corresponding Author.},
    Hesheng Wang\textsuperscript{\rm 2}
    }
\affiliations {
    \textsuperscript{\rm 1}MoE Key Lab of Artificial Intelligence, AI Institute, Shanghai Jiao Tong University, China\\
    \textsuperscript{\rm 2}Department of Automation, Shanghai Jiao Tong University, China\\
    \{xyz9911, pyy030406, liuzhesjtu, wanghesheng\}@sjtu.edu.cn
}

\begin{document}

\maketitle

\begin{abstract}
Large Language Models (LLMs) have demonstrated potential in Vision-and-Language Navigation (VLN) tasks, yet current applications face challenges. While LLMs excel in general conversation scenarios, they struggle with specialized navigation tasks, yielding suboptimal performance compared to specialized VLN models. We introduce FLAME (\textbf{FLAM}ingo-Architected \textbf{E}mbodied Agent), a novel Multimodal LLM-based agent and architecture designed for urban VLN tasks that efficiently handles multiple observations. Our approach implements a three-phase tuning technique for effective adaptation to navigation tasks, including single perception tuning for street view description, multiple perception tuning for route summarization, and end-to-end training on VLN datasets. The augmented datasets are synthesized automatically. Experimental results demonstrate FLAME's superiority over existing methods, surpassing state-of-the-art methods by a 7.3\% increase in task completion on Touchdown dataset. This work showcases the potential of Multimodal LLMs (MLLMs) in complex navigation tasks, representing an advancement towards applications of MLLMs in the field of embodied intelligence.
\end{abstract}

\begin{links}
\link{Code}{https://github.com/xyz9911/FLAME}
\end{links}

\section{Introduction}

\begin{figure}[t]
\centering
\includegraphics[scale=0.23]{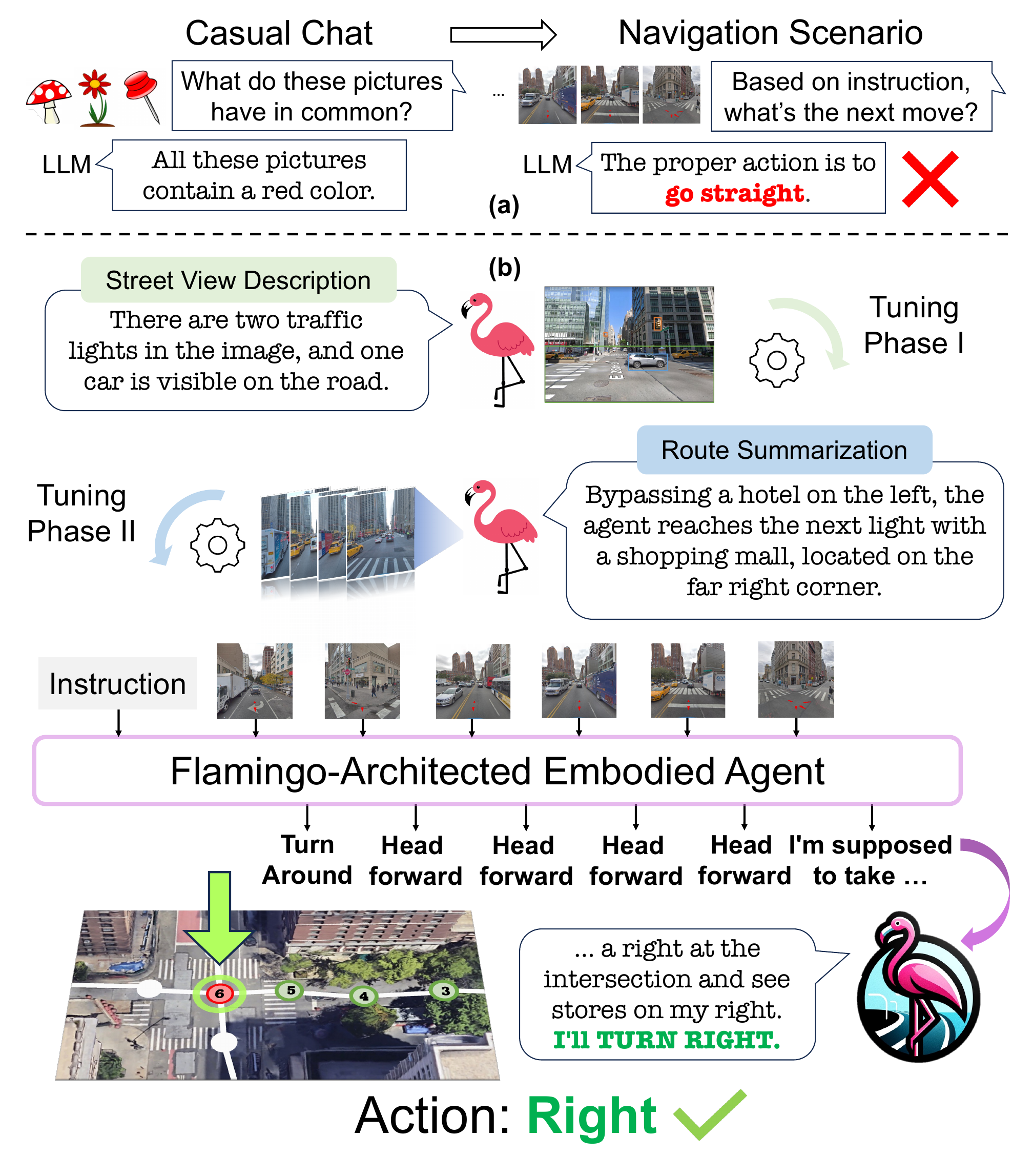}
\caption{LLM-based agents excel in conversation but often falter in specialized navigation tasks. Our agent, powered solely by a Multimodal LLM, demonstrates proficiency in navigation skills, efficiently adapting to navigation-specific scenarios through targeted finetuning phases.}
\label{fig:intro}
\end{figure}

Large Language Models (LLMs) \cite{openai2023gpt4, touvron2023llama1} have revolutionized the field of embodied intelligence. Vision-and-Language Navigation (VLN) \cite{2018r2r}, a fundamental task in embodied AI, challenges agents to navigate to a goal following human instructions in indoor or outdoor environments. This task demands sophisticated abilities in instruction comprehension, environmental understanding, and decision-making, which can be effectively managed by LLMs. Recent approaches have integrated LLMs into VLN methods, either by translating visual data into language \cite{zhou2023navgpt, qiao2023march_in_chat, chen2024mapgpt} or by employing Multimodal LLMs (MLLMs) \cite{zhang2024navid,zhou2024navgpt2} for environmental perception.

However, the incorporation of general-purpose LLMs in VLN still faces critical challenges, primarily stemming from their inherent limitations in navigation-specific scenarios, as \Cref{fig:intro}(a) depicts. For text-only LLMs, translating visual data into language using visual foundation models can lead to information loss \cite{zhou2023navgpt}, resulting in a performance gap compared to VLN-specialized models. For Multimodal LLMs, while partially addressing the limitations of text-only LLMs, they often struggle to interact within navigation-specific scenarios. Recent attempts to incorporate MLLMs for navigation may have constrained their capabilities by either using them as auxiliary components to traditional VLN models \cite{zhou2024navgpt2} or by processing observations through video tokens \cite{zhang2024navid}, which require multiple forward passes per trajectory and significant computational resources in training. These approaches may not efficiently adapt MLLMs to navigation-specific scenarios, which limits their ability to leverage inherent capabilities for handling interleaved textual and visual inputs and affects overall performance.

Moreover, the application of MLLM in urban VLN remains unexplored, despite its importance alongside indoor VLN. Outdoor navigation presents unique challenges for introducing MLLM, including longer trajectory lengths (up to 55 iterations) and increased difficulty (40\% lower success rate compared to indoor navigation tasks).

To address these challenges, we introduce FLAME (\textbf{FLAM}ingo-Architected  \textbf{E}mbodied Agent), the first MLLM-based agent designed for urban VLN tasks, as shown in \Cref{fig:intro}(b). Based on Flamingo \cite{alayrac2022flamingo}, FLAME operates autoregressively and efficiently handles multiple perceptions without increasing context length, ensuring efficiency in end-to-end training and inference. We propose a three-phase tuning technique to adapt Flamingo model \cite{alayrac2022flamingo} to navigation tasks using augmented data: 1) Single perception tuning: Learning to describe street views. 2) Multiple perception tuning: Learning to summarize agent navigation. 3) End-to-End training and evaluation on VLN datasets. To support the first two tuning phases, we utilize GPT-4 \cite{openai2023gpt4} to synthesize captions and route summaries for the Touchdown environment \cite{2019touchdown}. Additionally, we synthesize navigation rationales for urban VLN datasets \cite{2019touchdown,schumann2020map2seq} to validate FLAME's reasoning capability \cite{wei2022cot}.

Our agent achieves remarkable computational efficiency, completing the entire training process in just 14 hours on a single A100 GPU. Experimental results demonstrate FLAME's superiority over existing methods on two urban VLN datasets: Touchdown \cite{2019touchdown} and Map2seq \cite{schumann2020map2seq}. Our approach significantly outperforms current state-of-the-art (SOTA) methods by 7.3\% Task Completion (TC) in Touchdown and 3.74\% TC in Map2seq. Our work not only benefits the field of VLN but also showcases the potential of MLLMs in navigation within complex environments and tasks.

In summary, our contributions are threefold: 

\begin{itemize}
\item We introduce FLAME, to our knowledge, the first agent based on Multimodal LLM (MLLM) for urban Vision-and-Language Navigation (VLN) tasks.

\item We propose a tailored three-phase tuning technique for adapting Flamingo into navigation scenarios using synthetic data, fully unleashing MLLM's power.

\item Experiments show the superiority of FLAME over current SOTAs. FLAME's performance proves that MLLMs can significantly outperform specialized models, opening new avenues for research in embodied AI.

\end{itemize}

\section{Related Works}

\subsection{Vision-and-Language Navigation}
Vision-and-Language Navigation (VLN) \cite{2018r2r} encompasses indoor \cite{2020reverie, ku2020rxr} and outdoor scenarios \cite{2019touchdown, schumann2020map2seq}, with most advancements focusing on indoor environments. Traditional VLN agents often lack advanced decision-making skills, prompting the integration of Large Language Models (LLMs) \cite{lin2024console, schumann2023velma}, leveraging the reasoning \cite{chen2024mapgpt} and dialogue capabilities \cite{qiao2023march_in_chat, long2023discuss_before_moving} of LLMs. These approaches either convert visual data to text \cite{zhou2023navgpt} or employ Multimodal LLMs (MLLMs) with intensive training \cite{zhang2024navid, zheng2024navillm}. Given the relative underexploration of outdoor VLN, our work addresses this gap by introducing an effectively adapted MLLM-based agent for urban VLN tasks.

\subsection{Multimodal Large Language Models}
The emergence of Multimodal Large Language Models (MLLMs) \cite{liu2023llava, alayrac2022flamingo, awadalla2023openflamingo} has expanded their application to various tasks, including captioning \cite{li2023blip2} and general instruction following \cite{dai2023instructblip}. These models demonstrate multimodal reasoning skills \cite{lu2022mm_reason} in chat conversations, handling single-turn, in-context \cite{sun2024emu2-chat}, and interleaved text and image inputs \cite{laurenccon2024obelics}. MLLMs have benefited from vision-and-language tuning, enabling them to process diverse modalities \cite{nextgpt}. However, the general pretraining is insufficient for expert navigation tasks. Our work addresses this gap by adapting an MLLM from general scenarios to navigation through a specialized tuning approach.

\subsection{Data Augmentation in Vision-and-Language Navigation}
To overcome data scarcity in navigation, various data augmentation techniques \cite{zhao2021aug1, huang2019aug2} have been proposed. These include utilizing a speaker module \cite{fried2018speaker,dou2022foam} to generate synthetic instructions, leveraging multilingual data \cite{li2022clear}, incorporating counterfactual information \cite{parvaneh2020counterfactual2, fu2020counterfactual3}, and altering environments \cite{li2022envedit, liu2021envmix}. For outdoor VLN, previous studies have explored training agents on instructions with different styles \cite{zhu2020vln_transformer}, pretraining on auxiliary tasks \cite{armitage2023pmvln} and using driving videos \cite{li2024vlnvideo}. However, the use of auxiliary training data to tailor MLLMs for outdoor navigation remains largely unexplored. Our work addresses this gap by using synthetic data to adapt MLLMs for urban VLN tasks.

\section{Method}

\begin{figure*}[t]
  \centering
  \includegraphics[scale=0.45]{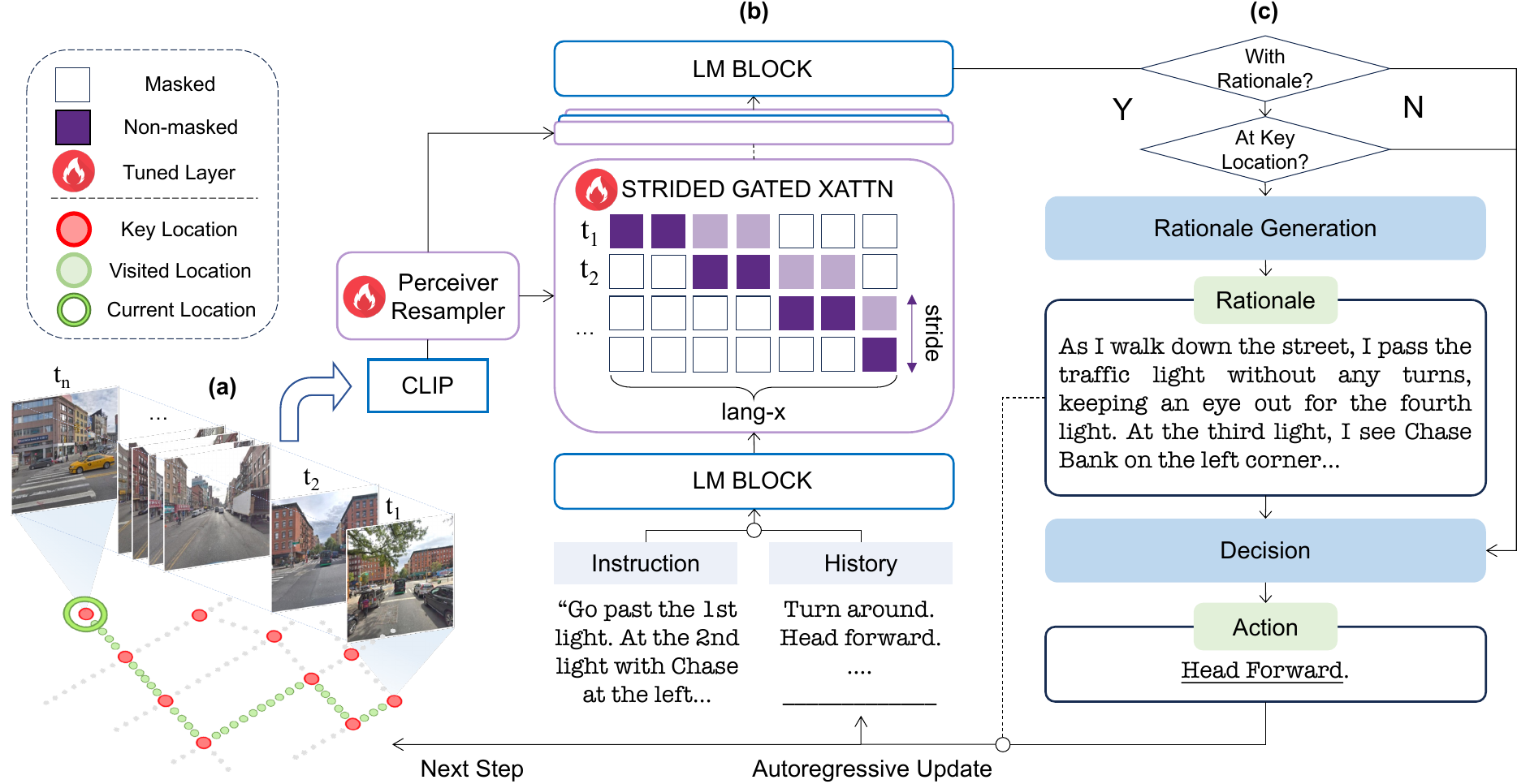}
  \caption{Overview of FLAME's navigation process at time step $t_n$. The architecture, based on Flamingo, integrates vision modules for observation processing and decoder blocks for instruction and history handling. The finetuned STRIDED GATED XATTN layers prioritize recent observations in cross-attention computation. At key locations (intersections), FLAME can engage in reasoning before decision-making or proceed directly to action selection. The navigation process is autoregressive.}
  \label{fig:method_1}
\end{figure*}

\begin{figure*}[t]
\centering
\includegraphics[scale=0.4]{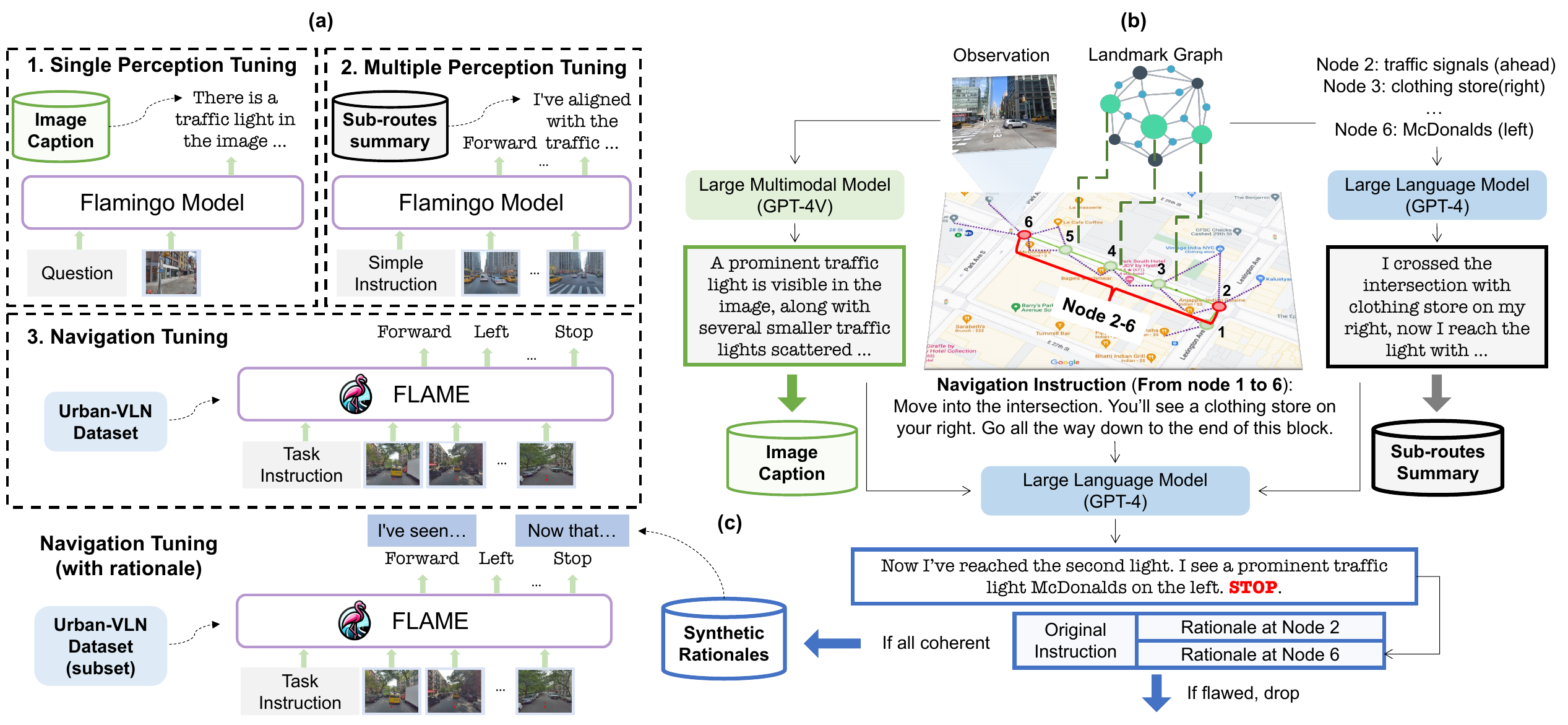}
\caption{Illustration of the three-phase tuning for navigation and synthetic data generation process. (a) The first phase trains the model on single-perception tasks. The second phase escalates to handling multi-perceptual input. Finally, the model undergoes an end-to-end finetuning. (b) We utilize LLMs to generate street view captions, route summaries and simple instructions to aid the training of the first two phase. (c) We further synthesize rationales to validate the reasoning capability of FLAME.}
\label{fig:method_2}
\end{figure*}

We present FLAME (Flamingo-Architected Embodied Agent), an agent for urban Vision-and-Language Navigation (VLN). Our method comprises three key components: 1) An architecture for urban navigation. 2) A three-phase tuning technique. 3) A synthetic data creation process.

\subsection{Task Formulation}
We formalize urban VLN as follows: Given a navigation instruction $I = \{w_1, w_2, ..., w_n\}$, an agent starts in an initial state $S_0$. At each timestep $t$, the agent selects an action $a_t \in \mathcal{A}$ based on the current observation $O_t$ and instruction $I$, where $\mathcal{A} = \{\text{Forward}, \text{Left}, \text{Right}, \text{Stop}, \text{Turn Around}\}$. The environment's state transition function $T: \mathcal{S} \times \mathcal{A} \rightarrow \mathcal{S}$ updates the agent's state to $S_{t+1}$. This process continues until the agent selects the \text{Stop} action. Navigation is successful if the agent stops at the target node or within one step of it. \Cref{fig:method_1}(a) provides an illustration of agent's observations and trajectory.

\subsection{FLAME Architecture}
FLAME builds upon the Flamingo architecture \cite{alayrac2022flamingo}, leveraging cross-attention to process visual and textual inputs without extending context length, while ensuring single-pass fine-tuning per trajectory. We introduce two key adaptations to Flamingo for urban VLN:

\subsubsection{Strided Cross-Attention}
To handle the large number of observations in urban VLN, we implement strided cross-attention \cite{child2019sparse_attention} in the cross attention layer, as depicted in \Cref{fig:method_1}(b). The Perceiver Resampler module transforms CLIP \cite{radford2021clip} features into a compact set of visual tokens, with $N_r$ tokens per observation. Let $X_t^v \in \mathbb{R}^{N \times d}$ represent flattened visual tokens up to timestep $t$, where $N = N_r \cdot t$. The LM block outputs a word embedding matrix $X$, with $X_t$ as the the segment of context corresponds to $t$, and $l$ as the stride length. Pattern $S = \{S_1, ..., S_t\}$ defines visual token indices for attention at $t$, with $S_t = \{k, k+1, ..., t \cdot N_r\}$ and $k = \max (0, (t - l) \cdot N_r)$. The strided cross-attention score is given by:

\begin{equation}
   A(X_t, S_t) = \text{CMA}(X_t W_Q, (x_j^v W_K)_{j \in S_t}, (x_j^v W_V)_{j \in S_t}),
\end{equation}

where CMA denotes cross-modal attention, \(W_Q\), \(W_K^v\), \(W_V^v\) are learnable parameters, and \(x_j^v\) refers to the \(j\)-th visual token. This approach is aimed at prioritizing recent observations, thereby augmenting the system's proficiency in identifying significant features in the dynamic environment.

\subsubsection{Action Prediction}
As \Cref{fig:method_1}(c) shows, FLAME predicts actions based on the instruction $I$, current observation $O_{t}$, history of observations $O_{\leq t-1}$ and previous actions. By default, the next action $a_t$ is obtained by:

\begin{equation}
    a_t = \text{MLLM}(I, O_1, a_1, ..., O_{t-1}, a_{t-1}, O_t).
\end{equation}

Optionally, to make agent's thought process transparent and understandable, FLAME generate rationales at key locations (i.e., intersections) before action prediction, similar to ReAct \cite{yao2022react}. Let $R_t$ denote the generated rationale, the next action can be obtained by:

\begin{equation}
    a_t = \text{MLLM}(I, O_1, a_1, ..., O_{t-1}, a_{t-1}, O_t, R_t).
\end{equation}

\subsection{Three-Phase Tuning for Navigation}
To adapt Flamingo for urban VLN tasks, we propose a three-phase tuning paradigm, as depicted in \Cref{fig:method_2}(a).

\subsubsection{Single Perception Tuning}
In the first phase, we train the model on a street view captioning task to strengthen its feature recognition abilities. Given a dataset $\mathcal{D}_{p1} = \{ \tau^{(i)} \}_{i=1}^N$, where $\tau^{(i)} = \{ (P^{(i)}, O^{(i)}, c^{(i)}) \}$ consists of a caption prompt $P^{(i)}$, an observation $O^{(i)}$, and a ground truth caption $c^{(i)}$, the training objective is:

\begin{equation}
    \mathcal{L}_{p1} = -\sum_{i=1}^N \log p(c^{(i)} | P^{(i)}, O^{(i)}; \theta),
\end{equation}
where \(\theta\) represents learnable parameters of the model.

\subsubsection{Multiple Perception Tuning}
The second phase focuses on synthesizing sequential observations and performing actions. Using a dataset $\mathcal{D}_{p2} = \{ \tau^{(i)} \}_{i=1}^N$, where each instance $\tau^{(i)} = \{ (I^{(i)}, O^{(i)}_1, a^{(i)}_1, ..., O^{(i)}_{T^{(i)}}, a^{(i)}_{T^{(i)}}, s^{(i)}) \}$ consists of a simple instruction $I^{(i)}$, observations $O^{(i)}_{\leq T^{(i)}}$, actions $a^{(i)}_{\leq T^{(i)}}$ and a ground truth route summary $s^{(i)}$. We supervise two objectives:

\begin{equation}
    \mathcal{L}_{sum}^{(i)} = \log p(s^{(i)} | I^{(i)}, O_{\leq T^{(i)}}^{(i)}; \theta),
\end{equation}

\begin{equation}
    \mathcal{L}_{act}^{(i)} = \sum_{t=1}^{T^{(i)}} \log p(a_t^{(i)} | I^{(i)}, O_{\leq t}^{(i)}, a_{\leq t-1}^{(i)}; \theta),
\end{equation}

where $\mathcal{L}_{sum}^{(i)}$ is the route summarization loss and $\mathcal{L}_{act}^{(i)}$ is the imitation loss. The total loss for the second phase is:

\begin{equation}
    \mathcal{L}_{p2} = -\sum_{i=1}^N \Big( \mathcal{L}_{act}^{(i)} + \mathcal{L}_{sum}^{(i)} \Big).
\end{equation}

\subsubsection{End-to-End Navigation Tuning}
Finally, FLAME is finetuned on a VLN dataset $\mathcal{D}_{nav} = \{ \tau^{(i)} \}_{i=1}^N$, where $\tau^{(i)} = \{ (I^{(i)}, O^{(i)}_1, a^{(i)}_1, ..., O^{(i)}_{T^{(i)}}, a^{(i)}_{T^{(i)}}) \}$. We minimize:
\begin{equation}
    \mathcal{L}_{nav} = -\sum_{i=1}^N \sum_{t=1}^{T^{(i)}} \log p(a^{(i)}_t | I^{(i)}, O^{(i)}_{\leq t}, a^{(i)}_{\leq t-1}; \theta).
\end{equation}

This multi-phased approach progressively builds the model's capabilities, from environment understanding to complex decision-making.

\subsection{Synthetic Data Generation}
\label{sec:synthetic_data}

To support the finetuning of our agent, we leverage LLMs to automatically synthesize street view captions, route summaries (\Cref{fig:method_2}(b)), and navigation rationales (\Cref{fig:method_2}(c)).

\subsubsection{Street View Caption Generation}
We focus on generating captions for street views at key locations, which are crucial for navigation. These images are processed by GPT-4V \cite{openai2023gpt4} to produce descriptions. To ensure diversity, we randomly select meta-prompts from a handcrafted prompt list, guiding the caption generation process. The prompt list contains specialized prompts refined manually to elicit diverse street view descriptions from GPT-4V.

\subsubsection{Route Summary Generation}
We construct a comprehensive knowledge graph of landmarks by combining places of interests from OpenStreetMap with additional map information, following the methodology of \cite{schumann2020map2seq}. Using this graph as input, we employ GPT-4 to generate detailed route summaries and simple navigation instructions between key locations.

\subsubsection{Rationale Generation for VLN Datasets}
To validate reasoning capabilities of our agent, we generate synthetic rationales for the VLN datasets. The process involves segmenting each VLN trajectory into sub-routes, then retrieving the corresponding image caption and summary for each route. We then utilize GPT-4 to generate a synthetic rationale at each key location, based on the retrieved information. To maintain data quality, we discard instances with flawed rationales. This results in a altered subset of the VLN dataset augmented with synthetic rationales, which facilitates end-to-end training with reasoning capabilities.

\begin{table*}[t]
    \centering
    \small
    \setlength{\tabcolsep}{4pt}
    \begin{tabular}{lccccccccccccccccc}
        \toprule
        & \multicolumn{7}{c}{Touchdown} && \multicolumn{7}{c}{Map2seq} \\
        \cmidrule{2-8} \cmidrule{10-16}
        & \multicolumn{3}{c}{Dev Set} && \multicolumn{3}{c}{Test Set} && \multicolumn{3}{c}{Dev Set} && \multicolumn{3}{c}{Test Set} \\
        \cmidrule{2-4} \cmidrule{6-8} \cmidrule{10-12} \cmidrule{14-16}
        Model & TC$\uparrow$ & SPD$\downarrow$ & nDTW$\uparrow$ && TC$\uparrow$ & SPD$\downarrow$ & nDTW$\uparrow$ && TC$\uparrow$ & SPD$\downarrow$ & nDTW$\uparrow$ && TC$\uparrow$ & SPD$\downarrow$ & nDTW$\uparrow$ \\
        \hline
        \midrule
        RCONCAT \shortcite{2019touchdown} & 10.60 & 20.4 & 22.50 && 11.80 & 20.40 & 22.90 && 17.10 & - & 30.70 && 14.70 & - & 27.70 \\
        GA \shortcite{2019touchdown} & 12.00 & 18.70 & 25.20 && 11.9 & 19.00 & 24.90 && 18.20 & - & 33.00 && 17.00 & - & 30.10 \\
        VLN-Trans \shortcite{zhu2020vln_transformer} & 15.00 & 20.30 & 27.00 && 16.20 & 20.80 & 27.80 && 18.60 & - & 31.10 && 17.00 & - & 29.50 \\
        ARC+L2S \shortcite{xiang2020arc} & 19.48 & 17.05 & - && 16.68 & 18.84 & - && - & - & - && - & - & - \\
        ORAR \shortcite{schumann2022orar} & 30.05 & 11.12 & 45.50 && 29.60 & 11.79 & 45.30 && 49.88 & \textbf{5.87} & 62.70 && 47.75 & 6.53 & 62.10 \\
        VELMA \shortcite{schumann2023velma} & 29.83 & 14.67 & 43.44 && 27.38 & 15.03 & 41.93 && 52.75 & 6.78 & 66.45 && 48.70 & 6.80 & 62.37 \\
        VLN-Video \shortcite{li2024vlnvideo} & 34.50 & 9.60 & - && 31.70 & 11.2 & - && - & - & - && - & - & - \\
        Loc4Plan \shortcite{tian2024loc4plan} & 34.50 & 10.50 & - && 32.90 & 11.50 & - && 48.00 & 7.00 & - && 45.30 & 7.20 & - \\
        \midrule

        RCONCAT* \shortcite{2019touchdown} & 9.27 & 19.14 & 16.40 && 8.34 & 21.02 & 15.70 && 8.25 & 38.71 & 12.51 && 7.63 & 41.22 & 12.03 \\
        GA* \shortcite{2019touchdown} & 7.12 & 20.91 & 13.27 && 6.63 & 21.76 & 12.44 && 6.50 & 38.74 & 11.67 && 6.07 & 41.25 & 10.88 \\
        ORAR* \shortcite{schumann2022orar} & 20.99 & 10.97 & 30.78 && 21.34 & 11.02 & 32.44 && 49.88 & 6.74 & 62.96 && 48.53 & 7.22 & 61.91 \\
        
        \midrule
        FLAME
         & \textbf{41.28} & \textbf{9.14} & \textbf{55.96} && \textbf{40.20} & \textbf{9.53} & \textbf{54.56}
        && \textbf{56.95} & 5.95 & \textbf{71.36} && \textbf{52.44} & \textbf{5.91} & \textbf{67.72} \\
        \bottomrule
    \end{tabular}
    \caption{Comparison with state-of-the-art models on Touchdown and Map2seq datasets. Models denoted by (*) utilize visual context consistent with FLAME's implementation. \textbf{Bold} values indicate best performance.}
    \label{tab:results}
\end{table*}

\section{Experiments}
\subsection{Experiment Setup}
\subsubsection{Datasets}
We evaluate our approach on two urban Vision-and-Language Navigation (VLN) datasets: Touchdown \cite{2019touchdown} and Map2seq \cite{schumann2020map2seq}, both set in the StreetLearn environment \cite{mirowski2018streetlearn}. Touchdown contains 9,326 instruction-trajectory pairs, while Map2seq comprises 7,672 pairs. The augmented dataset for the first two training phases contains 2,354 and 4,674 instances, respectively. Our agent is benchmarked against others on the original datasets.

We collect 6,518 (out of 9,326) and 6,291 (out of 7,672) pairs grounded with rationales at key locations using GPT-4 for Touchdown and Map2seq, respectively. We formulate synthetic pairs to create a dataset for evaluating reasoning capabilities. The reasoning performance is evaluated exclusively on the subset containing synthetic rationales to ensure a fair comparison.

\subsubsection{Metrics}
\label{sec:metrics}
For the original VLN datasets, we employ three metrics for performance evaluation: Task Completion (TC), Shortest-Path Distance (SPD) and Normalized Dynamic Time Warping (nDTW). Specifically, TC represents the percentage of success. SPD calculates the minimum distance from the stop location to the goal. nDTW assesses the overlap between the agent's and the ground truth trajectories. 

To further evaluate the agent's reasoning capabilities with synthetic rationales, we introduce two new metrics:
\begin{itemize}
    \item Rationale Coherence (RC):
    \begin{equation}
    \text{RC} = \frac{\sum_{i=1}^{N} \sum_{j=1}^{M_{i}} \text{CFR}_{\text{rc}}(I_{i}, \gamma(R_i^j), R_i^j)}{\sum_{i=1}^{N} M_{i}}
    \label{eq:rc}
    \end{equation}
    \item Rationale-Action Alignment (RA):
    \begin{equation}
        \text{RA} = \frac{\sum_{i=1}^{N} \sum_{j=1}^{K_{i}} \text{CFR}_{\text{ra}}(R_i^j, A_i^j)}{\sum_{i=1}^{N} K_{i}}
    \label{eq:ra}
    \end{equation}
\end{itemize}

Here, $I_i$ is the $i$-th instruction, $R_i^j$ is the agent's rationale, $A_i^j$ is the action, and $\gamma(R_i^j)$ is the ground truth rationale at the $j$-th key location. $M_i$ and $K_i$ are the number of key location overlaps and visited key locations, respectively. $N$ is the total number of instances. Functions $\text{CFR}_{\text{rc}}(\cdot)$ and $\text{CFR}_{\text{ra}}(\cdot)$ use GPT-4 \cite{openai2023gpt4} to evaluate rationale consistency and action alignment (true or false), respectively.

\subsubsection{Implementation Details}
\label{sec:implementation}
Our agent is built upon Otter and OpenFlamingo \cite{li2023otter, awadalla2023openflamingo}, integrating CLIP \cite{radford2021clip} and LLaMA \cite{touvron2023llama1}. To accommodate CLIP's input size, we crop and resize panoramas, which differs from the broader visual context used in existing models but aligns better with MLLM's nature. In the Touchdown task, we randomize the agent’s heading at the start of each trajectory by selecting one of the possible directions based on its neighboring nodes. Training for the first two phases takes 1 hour each, while the navigation fine-tuning requires 12 hours on a single A100 GPU.

\begin{table*}
    \centering
    \small 
    \setlength{\tabcolsep}{4pt}
    \begin{tabular}{lcccccccccccccccc}
        \toprule
               && \multicolumn{7}{c}{Touchdown (subset)} && \multicolumn{7}{c}{Map2seq (subset)} \\
    \cmidrule(lr){3-9} \cmidrule(lr){11-17}
        && \multicolumn{3}{c}{Dev Set} && \multicolumn{3}{c}{Test Set} && \multicolumn{3}{c}{Dev Set} && \multicolumn{3}{c}{Test Set} \\
 \cmidrule(lr){3-5} \cmidrule(lr){7-9} \cmidrule(lr){11-13} \cmidrule(lr){15-17}
        Param && TC$\uparrow$ & RC$\uparrow$ & RA$\uparrow$ && TC$\uparrow$ & RC$\uparrow$ & RA$\uparrow$ && TC$\uparrow$ & RC$\uparrow$ & RA$\uparrow$ && TC$\uparrow$ & RC$\uparrow$ & RA$\uparrow$ \\
        \hline
        \midrule
         ${T=0.0, P=1}$ && 37.45 & 84.73 & 97.41 && 37.73 & 85.96 & \textbf{99.44} && 49.42 & 82.54 & \textbf{97.28} && 45.60 & 84.38 & 96.61 \\
        \midrule
        ${T=0.7, P=4}$ && 38.27 & 82.07 & 96.62 && 36.77 & 82.37 & 97.40 && 50.19 & 83.97 & 97.07 && 48.92 & 85.86 & 96.21 \\
        ${T=1.0, P=4}$ && 37.27 & 80.90 & 97.47 && 36.96 & 84.12 & 97.77 && 46.33 & 81.34 & 96.63 && 46.58 & 86.97 & 97.74 \\
        \midrule
        ${T=0.7, P=8}$ && 37.18 & 84.83 & \textbf{98.60} && 37.54 & 83.71 & 97.71 && \textbf{52.32} & 82.38 & 96.92 && 45.60 & 85.32 & 97.97 \\
        ${T=1.0, P=8}$ && \textbf{39.17} & \textbf{86.48} & 97.48 && \textbf{37.82} & \textbf{86.80} & 96.91 && 48.84 & \textbf{86.61} & 96.98 && \textbf{49.51} & \textbf{88.00} & \textbf{99.25} \\
        \bottomrule
        \addlinespace
    \end{tabular}
    \caption{Performance comparison of the agent's reasoning capability on urban VLN dataset subsets under varying decoding temperatures and paths. The temperature $T$ regulates the randomness in sampling, and the number of decoding paths $P$ refer to the number of times the agent samples rationale-to-action pairs, with the final action determined through a voting process.}
    \label{tab:reasoning}
\end{table*}

\subsection{Comparison with SOTAs}
In the section, we compare FLAME with previous state-of-the-art (SOTA) approaches. As shown in \Cref{tab:results}, FLAME establishes new SOTA performance on both Touchdown and Map2seq datasets. On Touchdown test split, FLAME surpasses the previous SOTA, Loc4Plan \cite{tian2024loc4plan}, by 7.3\% in TC and 1.97\% in SPD. This demonstrates the superiority of our MLLM-based approach in comprehending navigational instructions and environmental cues, resulting in higher success rates and better path adherence. For Map2seq, FLAME outperforms VELMA \cite{schumann2023velma}, an LLM-based method, with a 3.74\% increase in TC and a 5.35\% improvement in nDTW. This highlights the advantage of Multimodal LLMs in capturing comprehensive information compared to text-only LLM.

Furthermore, we evaluated open-sourced methods in our visual setting, which features a restricted field of view similar to human vision. The results, marked with an asterisk (*) in \Cref{tab:results}, show a significant performance drop for baseline models, especially on the Touchdown dataset. This underscores the dependence of baseline methods on panoramic environmental perception. In contrast, FLAME excels at navigating with non-panoramic visual input.

\subsection{Reasoning Performance}
We evaluated FLAME's reasoning capabilities during navigation using the self-consistency approach \cite{wang2022self_consistency}, exploring various decoding paths and temperatures. The results are presented in \Cref{tab:reasoning}. Rationale Coherence (RC) and Rationale-Action Alignment (RA) consistently remained above 80\% and 95\% respectively, indicating robust rationale generation and strong consistency. Higher temperatures led to performance fluctuations, especially with fewer decoding paths. However, when we increase the number of decoding paths to 8, we see more pronounced improvements in both TC and RC. On Touchdown, FLAME achieved optimal TC performance with a temperature of 1.0 and 8 decoding paths, outperforming greedy decoding (T=0.0, P=1) by 1.72\% in TC and 1.75\% in RC. For Map2seq, with 8 decoding paths, FLAME surpassed greedy decoding by 3.91\% in TC and 3.62\% in RC on the test set. These results demonstrate that increased sampling diversity and a larger decoding budget enable FLAME to generate diverse rationales and effectively ensemble reasoning results, leading to improved decision-making. This provides strong evidence for the reasoning capabilities developed through synthetic rationale tuning and the advanced decision-making capacity of FLAME's architecture.

\begin{table}
    \centering
    \small
    \begin{tabular}{l|cc|cc}
        \toprule
        \multirow{2}{*}{Method} & \multicolumn{2}{c|}{Touchdown} & \multicolumn{2}{c}{Map2seq} \\
        & RC & RA & RC & RA \\
        \hline
        \midrule
        Human & 82.88 & 100.00 & 85.29 & 97.64 \\
        Vanilla & 92.79 & 99.10 & 88.82 & 95.88 \\
        \underline{Calibrated} & 81.08 & 100.00 & 84.12 & 97.06 \\
        \bottomrule
    \end{tabular}
    \caption{Comparison of metric calculation methods.}
    \label{tab:human}
\end{table}

\subsubsection{Metric Calculation}
To validate the reliability of our automatic metric calculations (Eqs. \ref{eq:rc} and \ref{eq:ra}), we conducted human evaluations on 50 instances each from the Touchdown and Map2seq datasets. The results, presented in \Cref{tab:human}, revealed discrepancies between the vanilla method and human assessments. However, our calibrated approach significantly reduced these disparities, demonstrating the human-comparable reliability of the automatic evaluation method. Consequently, we employ this calibrated evaluation technique as default. Specifically, We calibrate the outputs of $\text{CFR}_{\text{rc}}$ and $\text{CFR}_{\text{ra}}$ as follows:
\begin{itemize}
    \item When action at key location is incorrect: If $\text{CFR}_{\text{ra}} = 1$ but $\text{CFR}_{\text{rc}} = 1$, we force $\text{CFR}_{\text{rc}}$ to 0.
    \item When action at key location is correct: If $\text{CFR}_{\text{rc}} = 1$ but $\text{CFR}_{\text{ra}} = 0$, we force $\text{CFR}_{\text{ra}}$ to 1.
\end{itemize}

\subsection{Analyses}
\label{sec:analysis}
This section presents a comprehensive analysis of our approach, evaluating the strided cross-attention module, the three-phase tuning technique, and providing qualitative insights of navigation details.

\begin{figure}
\includegraphics[scale=0.16]{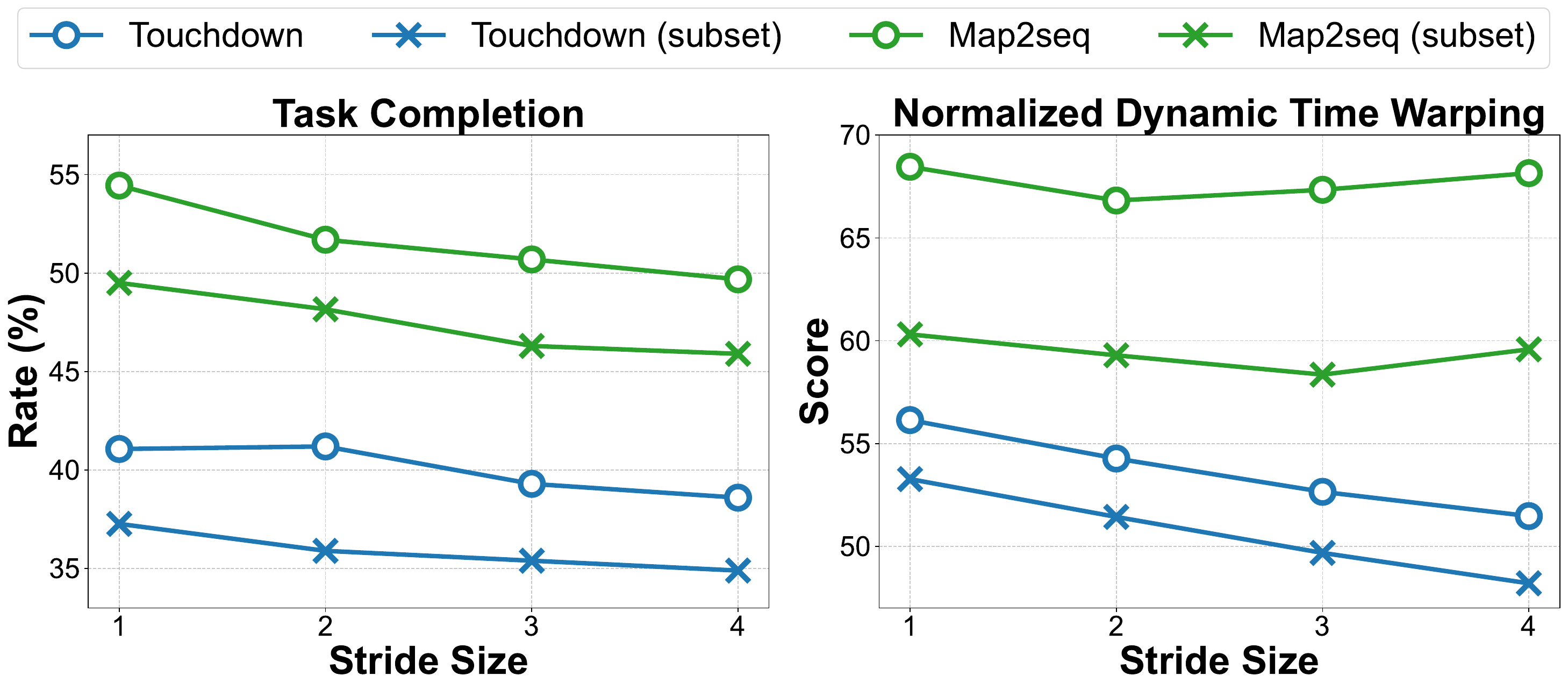}
\caption{The effect of varying strides on TC and nDTW.}
\label{fig:strides}
\end{figure}

\begin{figure*}[t]
\centering
\includegraphics[scale=0.48]{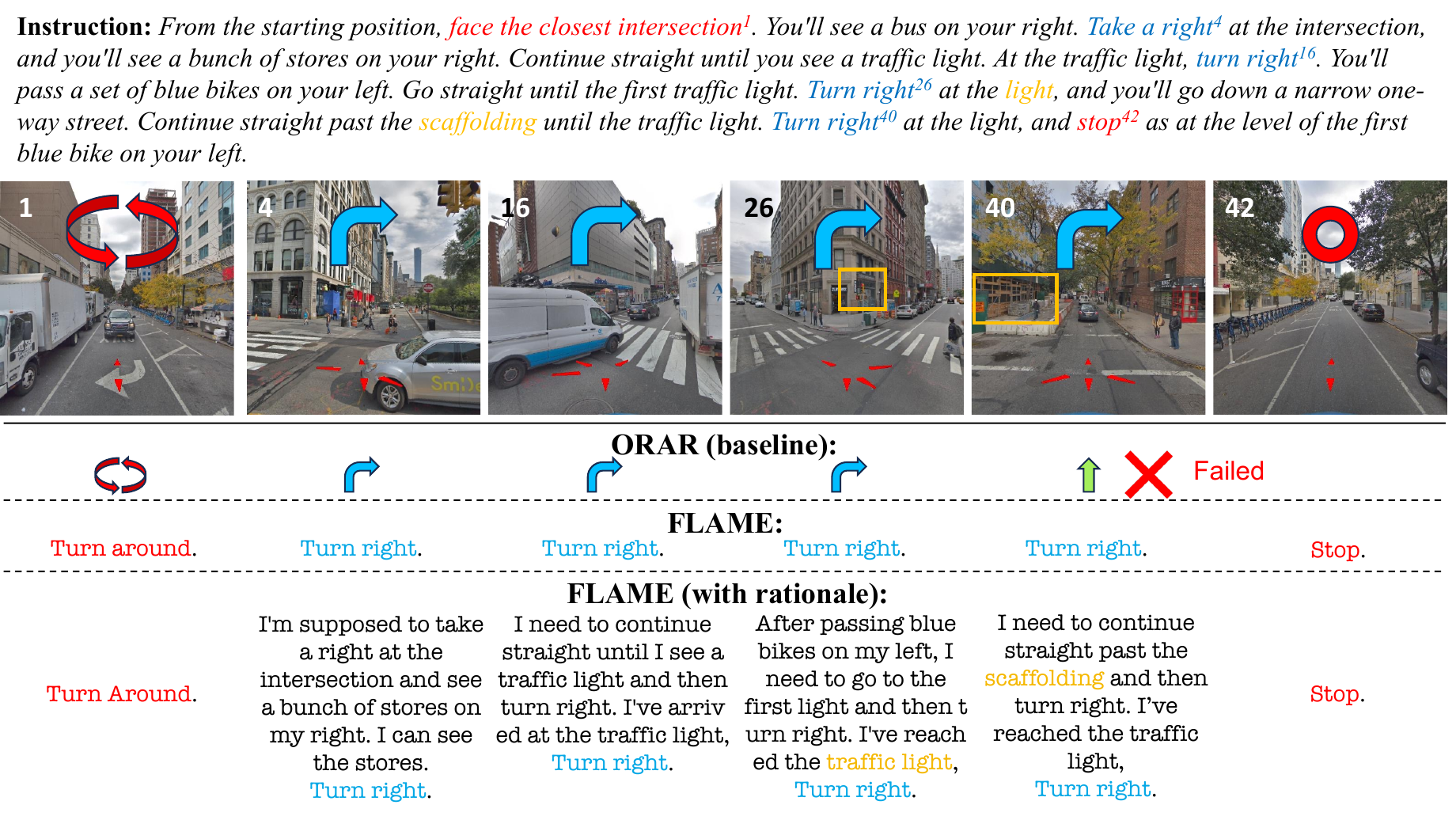}
\caption{Qualitative analysis of FLAME's navigation performance. Superscripts on words and numbers in the top left corner of each image indicate the count of viewpoints encountered by the agent. The ground-truth actions are represented by colored arrows: red circular arrows for turn around, blue for turn right, and a circle for stop. Keyword and landmark alignment is highlighted by matching colors in responses and instructions. The top row shows actions taken by the baseline method \cite{schumann2022orar}, while subsequent rows display FLAME's responses.}
\label{fig:demo}
\end{figure*}

\subsubsection{Effect of Strided Cross Attention}
To investigate the effect of the strided cross attention module, we conduct experiments on different stride sizes on both original datasets and subsets in \Cref{fig:strides}. We observe that increasing stride size generally correlates with a decrease in task completion rates, highlighting the importance of prioritizing current observations over longer history in decision-making processes. Paying full attention to numerous observations appears to overwhelm the agent, negatively impacting performance. Based on these findings, we set the default stride size to 1 in our implementation. However, nDTW scores for Map2seq fluctuate rather than decline consistently. This suggests stride size may be less critical for path-related performance in tasks less dependent on visual information. The lower TC on subset can be attributed to fewer training instances.

\subsubsection{Effectiveness of Three-Phase Tuning Technique}

\begin{table}[t]
    \centering
    \small
    \setlength{\tabcolsep}{4pt}
    \begin{tabular}{cc|ccc|ccc}
        \toprule
        \multirow{2}{*}{P1} & \multirow{2}{*}{P2} & \multicolumn{3}{c|}{Touchdown} & \multicolumn{3}{c}{Map2seq} \\
        && TC$\uparrow$ & SPD$\downarrow$ & nDTW$\uparrow$ & TC$\uparrow$ & SPD$\downarrow$ & nDTW$\uparrow$ \\
        \hline
        \midrule
        $\times$ & $\times$ & 40.16 & \textbf{9.02} & 55.31 & 53.57 & 6.55 & 66.70 \\
        $\checkmark$ & $\times$ & 41.15 & 9.19 & 55.26 & 54.69 & 6.10 & 69.70 \\
        $\checkmark$ & $\checkmark$ & \textbf{41.28} & 9.14 & \textbf{55.96} & \textbf{56.95} & \textbf{5.95} & \textbf{71.36} \\
        \bottomrule
    \end{tabular}
    \caption{Ablation study of three-phase tuning for navigation.}
    \label{tab:ablation-1}
\end{table}

\Cref{tab:ablation-1} presents our investigation into the impact of three-phase tuning on navigation performance. The first row, which corresponds to the vanilla MLLM checkpoint that bypassed the first (P1) and second (P2) phases, demonstrates sub-optimal navigation performance. Implementing the first phase tuning yields increases of 0.99\% and 1.12\% in TC for dev splits of Touchdown and Map2seq, respectively, suggesting improved environmental familiarity. The second phase of tuning leads to optimal performance across both datasets, indicating the importance of tuning with sequential observations. These results underscore the crucial role of phased learning in equipping the MLLM with advanced navigational skills, from single perception to sequential trajectory understanding, facilitating adaptation from general scenarios. Notably, even the non-phased training surpasses current state-of-the-art results, further validating the efficacy of FLAME's architecture.

\subsubsection{Qualitative Analysis of Navigation}
To illustrate the navigation capabilities of FLAME and compare it with the baseline method \cite{schumann2022orar}, we present a qualitative example in \Cref{fig:demo}. In this instance, ORAR fails at the fourth intersection, while FLAME successfully completes the navigation task. The agent's responses, accompanied by rationales, accurately identify key landmarks such as ``traffic light" and ``scaffolding", which align closely with the given instructions. FLAME demonstrates proficiency in following instructions and capturing salient environmental details. This example highlights the MLLM-based approach's effectiveness in correlating specific environmental features with verbal navigation instructions.

\section{Conclusion}
In this paper, we introduced FLAME, a Multimodal LLM-based agent for urban Vision-and-Language Navigation tasks. By adapting the architecture through a novel three-phase tuning technique and synthetic data, FLAME achieves state-of-the-art performance in urban VLN. The comparison results and reasoning performance demonstrate FLAME's superior ability to integrate verbal and environmental cues for decision-making. The effectiveness of our proposed tuning technique and other components is validated through comprehensive analyses. These findings highlight the potential of Multimodal LLMs in complex navigation tasks.

\section*{Acknowledgments}
This work was supported in part by the Natural Science Foundation of China under Grant 62303307, in part by National Key R\&D Program of China under Grant No.2023YFB4705700, in part by Shanghai Municipal Science and Technology Major Project under Grant 2021SHZDZX0102 and in part by the Fundamental Research Funds for the Central Universities.

\bibliography{aaai25}

\end{document}